\documentclass[conference]{IEEEtran}
\usepackage{times}

\usepackage[numbers]{natbib}
\usepackage{multicol}
\usepackage[bookmarks=true,pdfborder={0 0 0}]{hyperref}

\usepackage{graphicx}
\usepackage{amsmath}
\usepackage{amssymb}
\usepackage{algorithm}
\usepackage{algpseudocode}
\usepackage{adjustbox}
\usepackage{colortbl}
\usepackage{booktabs}
\usepackage{multirow}
\usepackage{makecell}
\usepackage[table]{xcolor}
\definecolor{baselinegray}{RGB}{246,246,246}
\usepackage{cuted}
\usepackage{caption}

\begin{document}

% paper title
\title{Learning When to Jump for Off-road Navigation}

% You will get a Paper-ID when submitting a pdf file to the conference system
% \author{Author Names Omitted for Anonymous Review. Paper-ID [197]}

\author{\authorblockN{Zhipeng Zhao, Taimeng Fu, Shaoshu Su, Qiwei Du,\\ Ehsan Tarkesh Esfahani,  Karthik Dantu, Souma Chowdhury, and Chen Wang}
\authorblockA{University at Buffalo, NY 14260, USA}}

\maketitle

\begin{strip}
  \vspace{-14mm}
  \centering
  \includegraphics[width=\textwidth]{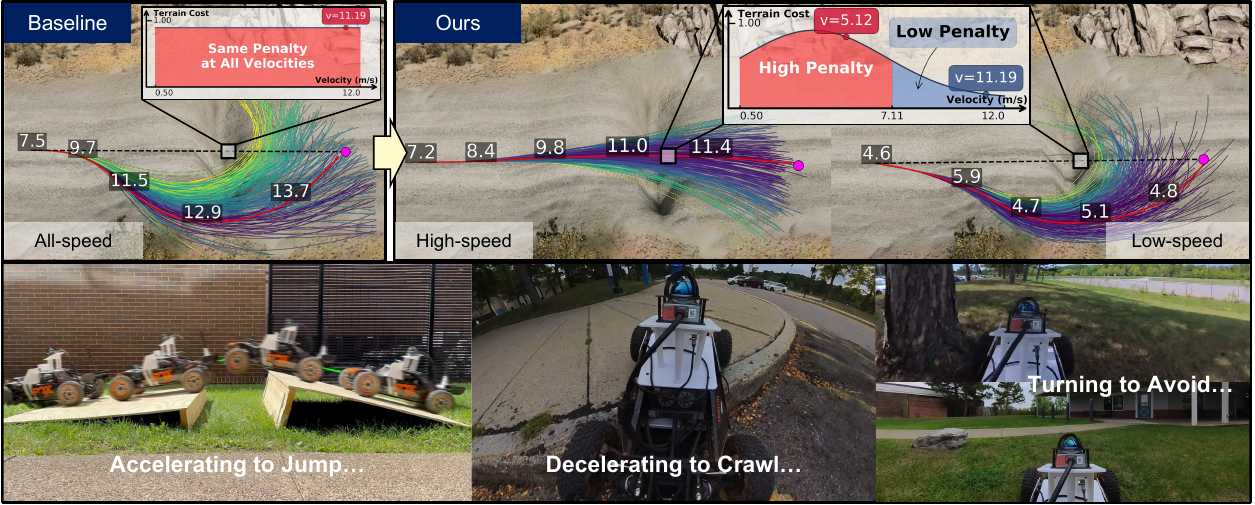}
  % \vspace{-4mm}
  \captionof{figure}{Challenging terrains often require motion strategies that span from aggressive maneuvers to cautious crawling. Our Motion-Aware Traversability (MAT) model enables the vehicle to decide when to accelerate to jump over a ditch, slow down to crawl over curbs, and steer to avoid obstacles. In contrast, baseline methods assign uniformly high costs to features like ditches, causing the vehicle to simply avoid them rather than selecting motion strategies that could traverse them more efficiently.}
  \vspace{-4mm}
  \label{fig:intro}
\end{strip}

\begin{abstract}
Low speed does not always guarantee safety in off-road driving. For instance, crossing a ditch may be risky at a low speed due to the risk of getting stuck, yet safe at a higher speed with a controlled, accelerated jump. Achieving such behavior requires path planning that explicitly models complex motion dynamics, whereas existing methods often neglect this aspect and plan solely based on positions or a fixed velocity. To address this gap, we introduce Motion-aware Traversability (MAT) representation to explicitly model terrain cost conditioned on actual robot motion. Instead of assigning a single scalar score for traversability, MAT models each terrain region as a truncated Gaussian function of velocity. During online planning, we decompose the terrain cost computation into two stages: (1) predict terrain-dependent Gaussian parameters from perception in a single forward pass, (2) efficiently update terrain costs for new velocities inferred from current dynamics by evaluating these functions without repeated inference. We develop a system that integrates MAT to enable agile off-road navigation and evaluate it in both simulated and real-world environments with various obstacles. Results show that MAT achieves real-time efficiency and enhances the performance of off-road navigation, reducing path detours by 75\% while maintaining safety across challenging terrains. Code is available at \textcolor{blue}{\href{https://github.com/sair-lab/mat}{https://github.com/sair-lab/mat}}.

\end{abstract}

\IEEEpeerreviewmaketitle

% \section{Introduction}
% This demo file is intended to serve as a ``starter file" for the
% Robotics: Science and Systems conference papers produced under \LaTeX\
% using IEEEtran.cls version 1.7a and later.  

\section{Introduction}
Off-road navigation presents unique challenges for autonomous robots \cite{borges2022survey, cai2025pietra, wang2025imperative}, where safety and efficiency depend not only on terrain geometry but also on the robot’s motion dynamics. For example, crossing a ditch at a moderate speed may lead to getting stuck at the bottom, but a higher, controlled speed can generate sufficient momentum to jump over it safely. Conversely, navigating bumpy roads and ruts may only be safe at low speeds to prevent instability. Such situations highlight that traversability is not a static property of terrain appearance, as it must also account for how the robot moves.

Most existing traversability estimation methods overlook this coupling between terrain and motion dynamics. 
They typically only infer position-dependent cost maps from terrain appearance and assign static scalar values to each location, independent of the robot’s intended motion.
These methods include analytical approaches based on geometric features \cite{krusi2017driving, dixit2021step} and learning-based models trained on various supervisions such as semantic labels \cite{meng2023terrainnet}, terrain elevation \cite{chen2024learning}, past vehicle-terrain interactions \cite{seo2023scate, seo2024metaverse, datar2025traverse}, and expert trajectories \cite{seo2023learning, jung2024v, triest2023learning}.
While effective for avoiding obstacles or following smooth trails, these methods share a fundamental limitation: they cannot distinguish situations where the same terrain can be safe or unsafe depending on the robot’s speed or motion profile.
In reality, traversability is motion‑dependent, and ignoring it yields conservative or suboptimal plans, which could be problematic in time‑critical missions \cite{pokhrel2025dom}.

Recent efforts have attempted to consider robot motion in terrain cost computation, but significant limitations remain.
Analytical formulations struggle to generalize, as precise physical models that capture velocity effects across varied terrains are difficult to derive.
For example, \citet{han2023model} estimate vehicle orientations from terrain elevations under a constant ground-contact assumption and combine them with velocities to compute traversability constraints via simplified physics. However, this assumption excludes agile maneuvers such as jumps, 
% and the terrain geometry is oversimplified to an orientation feature, 
and the terrain geometry is reduced to a slope-based feature, 
limiting its applicability to complex terrains.
From a learning perspective, HDIF \cite{castro2023does} jointly inputs the terrain patch and the robot’s velocity into the neural network.
However, since this requires repeated neural inferences for every velocity change, real-time cost updates during optimization are computationally infeasible.
% since each region would require repeated neural inferences for new velocities, updating the terrain cost for continuously changing speeds is computationally infeasible for real-time planning.
In practice, HDIF assumes a fixed velocity, failing to incorporate actual motion into traversability cost. Furthermore, learning the traversability-velocity relationship is difficult without providing labels covering the full velocity spectrum, and therefore, its predictions tend to be dominated by visual features.
% Velociraptor \cite{triest2024velociraptor} instead learns a speedmap from expert demonstrations and uses it as an auxiliary cost alongside the one from the traversability map. 
% While this helps constrain the speed, the traversability estimate remains independent of velocity. Therefore, difficult terrains are still penalized regardless of feasible motion strategies.
These limitations motivate a traversability representation that can efficiently condition terrain cost on both terrain perception and actual robot motion for real-time use.

To address this gap, we propose a Motion‑aware Traversability (MAT) representation that explicitly conditions terrain cost on the robot’s motion dynamics.
As illustrated in Fig. \ref{fig:intro}, each pixel is represented as a function of velocity rather than a scalar score. The function is modeled as a truncated Gaussian over a fixed velocity range, parameterized by terrain‑dependent attributes: peak difficulty, the velocity at which it occurs, and sensitivity to deviations. 
These compact parameters capture richer terrain characteristics than a single score.
% and avoid the need for exhaustive labeling or training across all possible velocities
Moreover, because these Gaussian functions can be reliably estimated from only a few trials, it enables the model to extrapolate traversability to velocities never observed during training, providing wider coverage over untested motion conditions. This property further strengthens MAT’s ability to generalize to out-of-distribution conditions.

We further develop an off-road navigation system by integrating MAT with a Model Predictive Path Integral (MPPI) \cite{williams2016aggressive} planner, enabling real-time adaptation of terrain costs to the motion inferred under system dynamics.
The computation is decomposed into two stages: (1) a single neural inference predicts Gaussian parameters for the entire map, and (2) as the motion changes, new terrain costs can be rapidly inferred by calling these functions without re-running the perception network.
This allows the robot to reason about when a dynamic maneuver, such as a jump, is safer and faster, ultimately planning more efficient and robust off-road trajectories.

Our main contributions can be summarized as:
\begin{itemize}
    \item We introduce a motion-aware traversability (MAT) representation that conditions terrain cost on both perception and motion, enabling robust and agile off-road planning.
    \item Based on MAT, we develop an efficient off-road navigation system that knows when to jump, enabling real-time motion-conditioned terrain cost updates and optimization.
    \item We validate the system in diverse simulated (BeamNG) and real-world environments with various obstacles, demonstrating its real-time efficiency and reduced traversal distance, time, and energy over the baseline.
\end{itemize}

\section{Related Works}
\label{sec:related_work}
Traversability quantifies the difficulty of moving through a region and serves as the terrain cost in off-road navigation.
Most methods infer a value per region solely from perception, resulting in position-only traversability.
Recent studies consider how motion affects difficulty and incorporate this into the traversability, which we term motion-enhanced traversability.
% Most methods infer a single scalar per region, regardless of the motion on it, which results in a terrain cost that depends only on position for planning. 
% We refer to these as \textbf{position-only traversability} methods.
% attempt to capture difficulty variations caused by motions and consider them in the terrain cost, which we term motion-enhanced traversability.

\subsection{Position-only Traversability}
Classical analytic approaches compute local geometric features such as roughness, slope, step height, and curvature from LiDAR scans and derive traversability from these statistics \cite{krusi2017driving,dixit2021step}.
However, LiDAR sparsity at high speeds and the lack of semantic context limit their applicability.
Learning‑based methods thus predict traversability from RGB/height images. These methods differ mainly in their supervision signals and learning strategies.
Some construct labels offline from point clouds and semantic information, then train image-based predictors \cite{meng2023terrainnet, chen2024learning}. 
For instance, TerrainNet \cite{meng2023terrainnet} maps image inputs to a multi-layer representation of elevation and semantic probabilities, which are then converted into terrain costs.

% TerrainNet \cite{meng2023terrainnet} proposes a multi-layer terrain representation containing min/max elevation and semantic probabilities for ground and ceiling layers. A network maps image inputs to multi-head outputs that are further converted into terrain cost.

To avoid manual annotation and hard-coded mapping from semantic to cost, self-supervised learning-based methods use vehicle-terrain interaction signals measured by onboard sensors (e.g., IMU, force/torque) as supervision. They derive labels such as vertical forces \cite{wellhausen2019should, seo2023scate, seo2024metaverse} or discrepancies between expected and actual pose/velocity \cite{datar2025traverse}, enabling automatic label generation from driving data.
Another direction leverages expert demonstrations as supervision.
In \cite{seo2023learning}, previously driven regions are labeled as traversable, and contrastive learning is used to generalize to visually similar but unlabeled regions. 
% Given that it covers only a small portion of all traversable regions, 
% contrastive learning is jointly applied to generalize to visually similar but unlabeled regions. 
V-STRONG \cite{jung2024v} extends it by incorporating SAM-based \cite{kirillov2023segment} segmentation masks in contrastive learning.
% performing contrastive learning with both expert trajectories and segmentation masks obtained from SAM \cite{kirillov2023segment}. 
In \cite{triest2023learning}, inverse reinforcement learning (IRL) is employed to learn a cost function, under which expert trajectories are optimal.

These methods produce position‑only traversability maps that ignore motion effect. 
% Such maps enable position optimization but cannot suggest how motion should adapt to terrain.
They can guide a robot toward flat regions and away from obstacles, but cannot tell how motion should adapt to specific terrain.
% reason about optimal motions over a specific terrain.
% to favor highly traversable regions (e.g., flat trails) and avoid challenging ones,
For example, they may assign a low cost to a mildly bumpy road, but can not suggest a reduced speed, or mark a ditch as untraversable though it could be crossed via a jump. 
This limitation leads to suboptimal plans.
% that could become problematic in complex environments without flat regions or urgent missions.

% which may be tolerable in simple scenarios with flat regions but problematic in complex environments or urgent tasks.

% Another line of work to address the issue exploits expert driving trajectories to supervise the learning.
% Considering the sparsity of LiDAR points during high-speed execution, which is insufficient for such analysis, and the absence of semantic knowledge, 

\begin{figure*}[th]
\label{sec:approach}
    \centering
    \includegraphics[width=\linewidth]{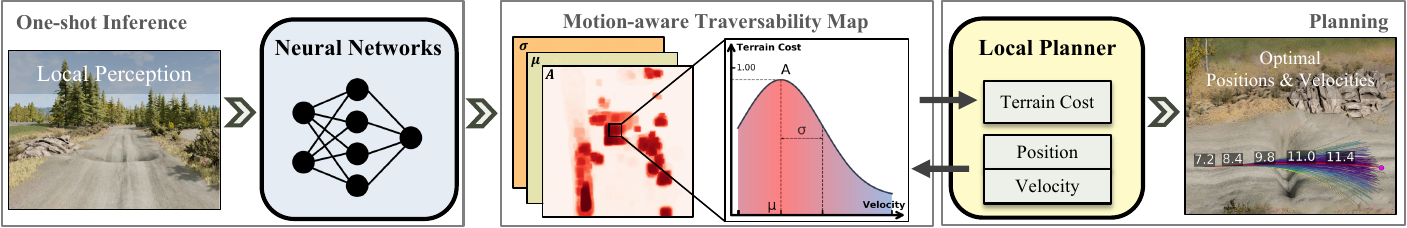}
    % \vspace{-5mm}
    \caption{The Architecture of the proposed pipeline. Given a local perception input (a LiDAR-based height map in practice), the neural network performs one-shot inference to generate the Motion-aware Traversability (MAT) map, where each cell encodes a Gaussian function of velocity.
    The local planner (e.g., MPPI) then samples trajectories based on vehicle dynamics, continuously querying the MAT map for terrain costs with position-velocity pairs to optimize motion in real time.} 
    \vspace{-2mm}
    \label{fig:network}
\end{figure*}

\subsection{Motion-enhanced Traversability}
% \vspace{-1mm}
A physics-based method \cite{han2023model} estimates vehicle orientations from the elevation map and combines them with vehicle velocities to calculate traversability constraints, aiming to prevent rollover and vehicle-ground separation.
However, its use of orientation-only terrain features fails to capture richer terrain geometry, and the assumption that the vehicle always adheres to the ground ignores agile strategies such as controlled jumps.
% Although velocity is considered, the method depends only on local pitch and roll features from the terrain and applies simplified physics formulations. 
% which overlooks detailed terrain geometry and limits applicability to complex terrains. 
% Moreover, the assumption that the vehicle always adheres to the ground 

Learning-based methods capture the motion effect directly from data. HDIF \cite{castro2023does} conditions its prediction on velocity by fusing Fourier-encoded speed signals with visual features.
% pairs each terrain patch with a velocity signal, labeled using the bandpower of the IMU’s z-axis linear acceleration.
% The velocity is Fourier-parameterized and fused with visual features from a CNN backbone to predict traversability.
Yet any change in velocity requires a new network inference, making real-time optimization computationally infeasible.
% Although the output depends on the velocity, the terrain cost for each region must be recomputed through new neural inferences whenever the velocity changes, 
As a result, HDIF uses a fixed velocity during planning.
Moreover, learning a robust traversability-velocity relationship is challenging without dense labels across speeds, and end-to-end predictions often become dominated by visual features.
Alternatively, Velociraptor \cite{triest2024velociraptor} and Salon \cite{sivaprakasam2025salon} learn speed maps \cite{yang2023learning, yoon2025state} as auxiliary cost.
They encourage adherence to expert velocities, such as high speeds on flat trails and lower speeds in vegetation, but only for regions that are deemed traversable by the traversability map. Since traversability remains motion-independent, terrains requiring specific motion strategies, such as jumping a ditch, are still penalized as untraversable. Besides, it fails to account for terrain-specific sensitivity, where a wide range of speeds might be safe.
% 
% Velociraptor \cite{triest2024velociraptor} produces both traversability and speed maps, which are supervised by expert driving speeds. 
% The speed map encourages small speed deviations from expert driving speeds for regions deemed traversable by the traversability map: high speed on clear trails and lower speed in vegetation. 
% But terrains that demand specific motion strategies (e.g., jumping a ditch) are still penalized as untraversable since the traversability map remains motion‑independent.

% However, its traversability output remains motion‑independent, so terrains that demand specific motion strategies (e.g., jumping a ditch) are still penalized as untraversable.
% For instance, it encourages high speed on clear trails and lower speed in tall vegetation when they are deemed traversable by the traversability map.
% leverages features from visual foundation models to

Beyond traversability estimation, a line of work advances agile motion by extending vehicle kinodynamics \cite{lee2023learning, datar2024learning, datar2024terrain, pokhrel2025dom, 2025vertiformer, zhao2024physord, fu2025anynav}. PHLI \cite{pokhrel2025dom} studies jumping behavior through in-air attitude control. It enables reorientation during airtime for safe landing and highlights the potential of aerial maneuvers for ground vehicles, but does not reason about when or where jumping is feasible, nor does it adapt to terrain geometry.

To handle complex off‑road scenarios, we need a representation that conditions terrain cost on both perception and robot motion, while allowing efficient and reliable online evaluation.

\section{Approach}
\subsection{Overview}
\begin{figure}[h]
  \centering
  \includegraphics[width=1.0\linewidth]{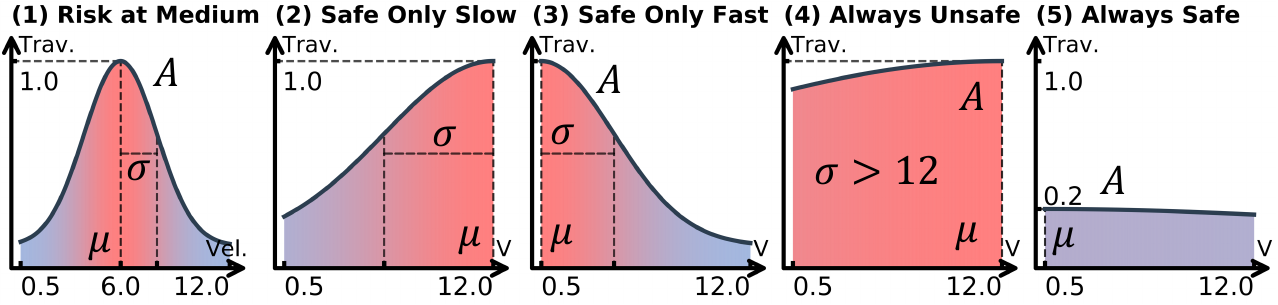}
  \caption{Representative Gaussian-modeled profiles for common terrains: (1) long trenches, (2) rocky terrain/bumps, (3) narrow ditches/gaps, (4) trees/boulders, (5) smooth dirt trails.}
  \label{fig:gaussian_assump}
  % \vspace{-2mm}
\end{figure}

A key challenge in off-road navigation is that traversability does not simply increase or decrease with speed: depending on the terrain, a moderate velocity can be hazardous while higher or lower speeds may ensure safety. To identify this property, a robot must model the traversability-velocity relationship and can extrapolate to unobserved velocities. Real datasets rarely contain dense velocity coverage, so directly regressing traversability from terrain-velocity input tends to fail on out-of-distribution velocities. To address this, we assume that traversability is a truncated Gaussian-shaped function of velocity. 
Crucially, since robot velocity is bounded by physical limits, this function is defined over a fixed range, allowing it to capture diverse terrain profiles including non-monotonic and monotonic risks, as illustrated in Fig. \ref{fig:gaussian_assump}. 
This assumption is also motivated by its ability to yield smooth, physically plausible extrapolations from only a few samples, and its empirical effectiveness across diverse off-road scenarios. With this assumption, we introduce the Motion-Aware Traversability (MAT) map that encodes this function at each terrain location.

The following sections describe: (1) how the MAT map is learned in a self-supervised manner from vehicle-terrain interaction, and (2) how a complete navigation system is built around the MAT map to enable real-time motion optimization.

\subsection{Motion-aware Traversability Learning}
The core idea of MAT is to predict \emph{terrain-dependent Gaussian parameters} from perception rather than directly regressing the final traversability cost. 
Our neural model outputs a three-channel map where each position $\mathbf x$ is represented by: peak penalty $A(\mathbf x)$, the velocity at which peak penalty occurs $\mu(\mathbf x)$, and the sensitivity to velocity deviations $\sigma(\mathbf x)$. Together, these parameters define a \emph{velocity-conditioned terrain cost function} for each $\mathbf x$.
This formulation captures more terrain properties, such as 
worst-case speed and sensitivity to deviations, instead of a single score that assumes uniform cost across speeds.
Besides, by learning function parameters rather than final cost under different velocities, MAT enables efficient evaluation of terrain cost at arbitrary velocities without extra neural inference, and provides reliable extrapolation to unseen velocities.

Concretely, given a height map $\mathbf{H} \in \mathbb{R}^{H \times W}$, we aim to learn a network $g_\theta$ that predicts a MAT map $\hat{\boldsymbol \Psi} \in \mathbb{R}^{H\times W\times3}$. 
For each position $\mathbf x$, the MAT map stores a vector of Gaussian parameters $\boldsymbol{\psi}(\mathbf{x})=[A(\mathbf{x}),\,\mu(\mathbf{x}),\,\sigma(\mathbf{x})]^\top$, which jointly define a terrain traversability cost function over velocity:
\begin{equation}
\label{eq:terrain_cost}
T(\mathbf{x}, v)\triangleq
T\!\big(v; \boldsymbol{\psi}(\mathbf{x})\big)
=
A(\mathbf{x})
\exp\!\left(
-\frac{(v-\mu(\mathbf{x}))^2}{2\,\sigma(\mathbf{x})^2}
\right).
\end{equation}
To learn $g_\theta$, we first construct self-supervised labels $\boldsymbol\Psi$ from repeated vehicle-terrain interactions, and then train an image-to-image network with a masked regression objective.

% The perception input is a bird’s-eye-view (BEV) height map $\mathbf{H} \in \mathbb{R}^{H \times W}$ constructed from accumulated LiDAR scans. It provides rich features (e.g., slopes, gaps) while remaining lightweight for real-time use. 

\subsubsection{Self-supervised Label Construction}
To create the MAT map label in the form of Gaussian parameters, we must observe how traversability changes when the same terrain is traversed at different velocities. 
We therefore execute $R$ runs over the same terrain with commanded speeds spanning $[v_{\min},v_{\max}]$. Each trajectory is resampled to a common set of positions $\{\mathbf{p}_i\}_{i=0}^{M}$ \cite{wang2023pypose, zhan2023pypose}. At each $\mathbf{p}_i$, we record the velocity $v_{r, i}$ and vehicle response in each run $r$, from which we derive the traversability cost $t_{r,i}$.
Given a small set of velocity-traversability samples $\mathcal{S}_i=\{(v_{r,i},\,t_{r,i})\}_{r=1}^R$, we estimate the Gaussian parameters $\boldsymbol{\psi}(\mathbf{p}_i)$ by fitting the model in \eqref{eq:terrain_cost}:
\begin{equation}
\boldsymbol{\psi}(\mathbf{p}_i)
=
\arg\min_{\boldsymbol{\psi}}
\sum_{r=1}^{R}
\Big[t_{r,i}-T\!\big(v_{r,i}; \boldsymbol{\psi}\big)\Big]^2.
\end{equation}
Here, sparse velocity observations suffice to recover a reliable traversability function over full velocities and provide effective supervision to the network via estimated parameters.

To quantify traversability cost $t_{r,i}$ at position $\mathbf{p}_i$ and velocity $v_{r,i}$, we derive it from vehicle-terrain interaction data rather than manual annotation. 
The intuition is that true traversability can only be revealed through real experience: visual appearance alone is insufficient to determine whether a rough-looking terrain is actually traversable.
We define traversability by jointly considering \textit{safety} and \textit{effort}, which are detailed below.

% To observe how traversability varies with velocity, we execute $R$ runs over the same uneven terrain with commanded speeds spanning $[v_{\min},v_{\max}]$.

% Each trajectory is resampled to a common set of positions $\{\mathbf{p}_i\}_{i=0}^{M}$. At each $\mathbf{p}_i$, we collect vehicle responses under different observed velocities $v_{r, i}$ from each run $r$.

% For each $(\mathbf{p}_i, v_{r, i})$, 
% with rollover risk, and time and energy consumption.
\paragraph{Safety (Rollover Risk)}
To capture the rollover risk under particular terrain-velocity combinations, we quantify it by how vehicle instability is amplified, rather than simply penalizing terrain unevenness as in prior methods \cite{datar2025traverse,fu2025anynav, meng2023terrainnet}.
A terrain-velocity pair is considered traversable if the vehicle either adheres closely to the terrain geometry or maintains smoother roll and pitch dynamics (e.g., a stable airborne transition). Conversely, it is deemed risky when vehicle roll or pitch angles are significantly amplified relative to the reference static orientation.
We estimate vehicle static orientation (roll, pitch) from wheel contact elevations using a quasi-static model similar to \cite{datar2025traverse}. Safe orientation bounds $\underline{\boldsymbol{\theta}}$ and $\overline{\boldsymbol{\theta}}$ are defined from the minimum and maximum static orientation angles, and the rollover risk is measured as the deviation of the vehicle's actual orientation $\boldsymbol{\theta}_{r, i}$ from these bounds:
\[
t_{r,i}^{\text{roll-raw}} =
\max\!\left(
0,\,
\|\boldsymbol{\theta}_{r, i} - \overline{\boldsymbol{\theta}}\|_\infty,\,
\|\underline{\boldsymbol{\theta}} - \boldsymbol{\theta}_{r, i}\|_\infty
\right),
\]
where $\|\cdot\|_\infty$ denotes the maximum deviation across roll and pitch. 
To account for delayed rollover effects (e.g., post-jump instability), we aggregate future instantaneous risks using exponential discounted factors $\gamma\!\in\!(0,1)$. The accumulated value is then normalized to $[0,1]$, with deviations of $90^\circ$ or greater mapped to $1$, yielding the final safety metric $t_{r, i}^{\text{roll}}$. 
% with $1$ representing the maximum allowable angular deviation ($90^\circ$)

\paragraph{Effort (Time and Energy)}
A region may be traversable with negligible rollover risk but require varying effort at different speeds.
For example, a ditch can be crossed either slowly or via a high-speed jump, but a low speed takes longer and can consume more energy.
To quantify this effort, we measure: (i) the travel time required to reach the goal, and (ii) the control energy, computed as the time integral of the throttle command from $\mathbf{p}_i$ to the goal.
Since absolute values depend on the distance to the goal, we normalize each quantity across all runs at the same $\mathbf{p}_i$. This yields effort terms $t_{r,i}^{\text{time}}$ and $t_{r,i}^{\text{energy}}$, which are directly comparable across velocities.
% effort using the remaining time and control energy (throttle integration over time) it takes to reach the goal.
% Both quantities are min-max normalized across runs at the same $\mathbf{p}_i$, producing $t^{\text{time}}_{r, i}$, $t^{\text{energy}}_{r, i}$.
The resulting self-supervised traversability score at position $\mathbf{p}_i$ under velocity $v_{r,i}$ is defined as:
\begin{equation}
    t_{r,i}=w_r\,t_{r,i}^{\text{roll}}
        +w_t\, t_{r,i}^{\text{time}}
        +w_e\, t_{r,i}^{\text{energy}},
\end{equation}
with $w_r=0.8$, $w_t=0.1$, $w_e=0.1$ in our implementation.

% \subsubsection{Velocity-dependent Traversability Modeling}
% For each position $\mathbf{p}_i$, we collect samples $\mathrm{S}_i=\{(v_{r,i},\,t_{r,i})\}_{r=1}^R$ from all runs.
% The resulting spatial profile $\boldsymbol{\psi}_i$ characterizes how traversability varies with velocity at each terrain location $\mathbf{p}_i$.
% For each position $\mathbf{p}_i$, the Gaussian parameters $\boldsymbol{\psi}(\mathbf{p}_i)$ are obstained with the $\mathcal{S}_i$ as described above.

After obtaining $\boldsymbol{\psi}(\mathbf{p}_i)$ at $\mathbf{p}_i$, we assign it to the corresponding pixel and its surrounding patch on the height map. Repeating this process over all positions yields a ground-truth MAT map $\boldsymbol \Psi \in \mathbb{R}^{H\times W\times3}$ for $\mathbf{H}$.
Since the label cannot cover the entire $\mathbf{H}$, we also generate a binary mask $\mathbf{M} \in \mathbb{R}^{H \times W}$ that indicates labeled pixels to enable image-to-image training.
% into the height map frame, we assign $\boldsymbol{\psi}(\mathbf{p}_i)$ 

\subsubsection{Neural Network}
To capture both local and global terrain geometry from $\mathbf{H}$, we utilize a U-Net-based \cite{ronneberger2015u} neural network as $g_\theta$.
% to predict the MAT map from $\mathbf{H}$
The encoder extracts multi-scale geometric cues, and the decoder upsamples them to the original resolution to produce per-pixel outputs. To ensure valid physical ranges, we apply per-channel sigmoid activations followed by affine scaling.
The network is trained using a masked mean-squared error loss computed over labeled pixels:
\begin{equation}
\mathcal{L}(\theta)
=
\frac{1}{\sum_{\mathbf{p}} \mathbf{M}(\mathbf{p}) + \varepsilon}
\sum_{\mathbf{p}} \mathbf{M}(\mathbf{p})
\big\|
g_\theta(\mathbf{H})(\mathbf{p}) - \boldsymbol{\Psi}(\mathbf{p})
\big\|_2^2.
\label{eq:masked-mse}
\end{equation}
Random spatial augmentations are applied during training to improve robustness.
After training, the neural model learns to associate geometric terrain patterns with Gaussian parameters that characterize how traversability varies with speed.
The resulting MAT map forms the basis for efficient, velocity-aware terrain cost evaluation during navigation.

\begin{figure}[t]
    \centering
    \includegraphics[width=\linewidth]{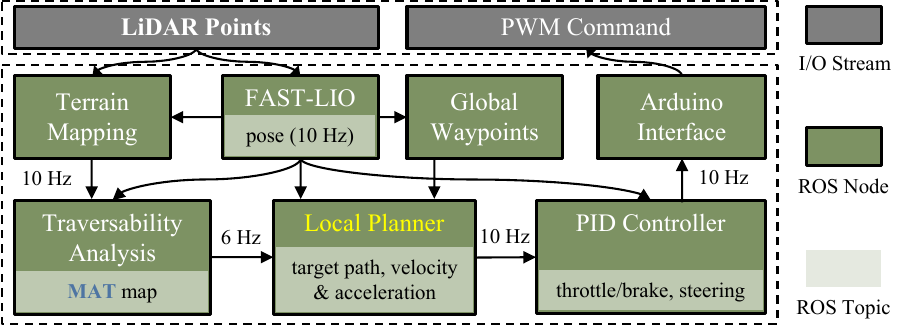}
    % \vspace{-6mm}
    \caption{Navigation system diagram showing data flow from LiDAR sensing to actuator commands, with ROS nodes performing perception, MAT estimation, planning, and control.} 
    \vspace{-2mm}
    \label{fig:system}
\end{figure}

\subsection{Motion-aware Navigation System}
\label{sec:system}
We develop a complete navigation system that integrates MAT with perception, planning, and control modules for agile off-road driving. As illustrated in Fig. \ref{fig:system}, the system ensures real-time motion planning when incorporating the MAT map, making motion-conditioned terrain cost optimization feasible on an edge device.
The system is implemented in ROS \cite{quigley2009ros}, with each component running as an independent node. It receives LiDAR packets and outputs Pulse-Width Modulation (PWM) commands for throttle and steering.

The perception begins with FAST-LIO \cite{xu2022fast}, providing LiDAR-inertial odometry at 10 Hz. 
A terrain mapping node, implemented following \cite{cao2022autonomous}, fuses recent registered scans to maintain a dense local point cloud. Our traversability analysis node then converts this local map into a 2D elevation map $\mathbf{H}$ and predicts the three-channel MAT map in a single inference.

The planning pipeline is initiated by the MAT map and the goal location $\mathbf g$ provided by a global waypoint node.
Unlike prior systems that optimize only positions to minimize terrain cost, we jointly optimize both positions and velocities to produce terrain-adaptive motion strategies.
This is enabled by integrating MAT such that standard local planners can evaluate terrain cost analytically for any new velocity.
We demonstrate this using an MPPI planner, widely adopted in off-road navigation, as shown in Algorithm~\ref{alg:mppi_mat}.
At each iteration, \(N\) control sequences \(\{\mathbf{u}_{0:T-1}^{(n)}\}_{n=1}^N\) are sampled around a nominal sequence and propagated through the vehicle dynamics:
\begin{equation}
\dot{\mathbf{x}} = v
\begin{bmatrix}
\cos \psi \
\sin \psi
\end{bmatrix},
\quad
\dot{\psi} = v\kappa,\quad
\dot{v} = a,\quad
\dot{\kappa} = u_\kappa,
\end{equation}
where the control inputs \(\mathbf{u}=[a,\,u_\kappa]\) are acceleration and curvature rate, and the state \(\mathbf{s}=[\mathbf{x},\,\psi,\,v,\,\kappa]\) includes position, heading, speed, curvature.
Each trajectory is evaluated using:
\begin{equation}
\mathbb{J} = \sum_{t=0}^{T-1}\Big[\mathbf{u}_t^\top \mathbf{R}\,\mathbf{u}_t
+\,T(\mathbf{x}_t,v_t)
+\,c_{\text{aux}}(\mathbf{s}_t, \mathbf  g)\Big] + \phi(\mathbf{s}_T, \mathbf g),
\end{equation}
where $\mathbf{R}$ penalizes control effort, $c_{\text{aux}}(\cdot)$ handles goal tracking and stability, and $\phi(\cdot)$ is the terminal cost.
The terrain cost $T(\mathbf{x}_t, v_t)$ is analytically evaluated with Gaussian parameters $\boldsymbol{\psi}(\mathbf{x}_t)=[A(\mathbf{x}_t),\,\mu(\mathbf{x}_t),\,\sigma(\mathbf{x}_t)]^\top$ available for $\mathbf x_t$ as in \eqref{eq:terrain_cost}.
Because $T(\mathbf{x}_t, v_t)$ is a closed-form function of velocity, it can be rapidly updated for a new velocity sampled during the iterative optimization process.
Given that MAT provides the parameters for the entire local environment, the planner can efficiently evaluate any \((\mathbf{x}_t, v_t)\) pairs sampled across all iterations.
After $K$ iterations, the optimized path $\mathbf{x}_{0:T-1}^*$ with velocity and control profile is obtained, specifying both where to go and how aggressively to travel.

A low-level PID controller executes the plan: a look‑ahead point on the optimized path determines the steering command, while the desired velocity and acceleration yield the throttle/brake command. 
An interface node translates these into PWM values and sends them over the serial port.

Conventional off-road navigation pipelines rely on position-only terrain cost $T(\mathbf{x}_t)$, which ignore the coupling between traversability and velocity and tend to produce suboptimal behaviors.
In contrast, MAT provides a velocity-conditioned terrain cost $T(\mathbf{x}_t, v_t)$ that enables the planner to reason about how speed affects risk.
The key contribution of our system is that it enables this richer, motion-aware traversability representation without increasing computational cost, ensuring real-time performance on resource-constrained platforms.

% \begingroup
% \setlength{\textfloatsep}{0pt}
% \begin{algorithm}[t]
% \caption{MPPI with Motion-aware Traversability (MAT)}
% \label{alg:mppi_mat}
% \begin{algorithmic}[1]
% \State $\textbf{Input:}$ initial state $\mathbf{s}_0$, nominal control sequence $\mathbf{u}_{0:T-1}$, elevation map $\mathbf{H}$, goal $\mathbf g$
% \State Predict the MAT map $(A, \mu, \sigma) = g_\theta(\mathbf{H})$
% \For{iteration $k = 1$ to $K$} 
% % \COMMENT{MPPI iterations}
%     \For{each sampled control perturbation $\delta \mathbf{u}_{0:T-1}^{(n)}$}
%         % \STATE  $\mathbf{x}_{0:T}^{(n)}$ under dynamics $F$
%         \For{$t = 1$ to $T-1$}
%             % \STATE $v_t^{(n)} \leftarrow \|\dot{\mathbf{x}}_t^{(n)}\|$
%             \State Propagate $\mathbf s_{t}^{(n)} \leftarrow F(\mathbf s_{t-1}^{(n)}, \mathbf u_{t-1}^{(n)})$
%             \State Position and velocity $\mathbf x_t^{(n)}, v_t^{(n)} \leftarrow \mathbf s_{t}^{(n)}$
%             \State Evaluate terrain cost $T_t^{(n)} \leftarrow
%             T(\mathbf x_t^{(n)}, v_t^{(n)}) $
%             % = A(\mathbf{x}_t^{(n)}) \exp[-(v_t^{(n)} -\mu(\mathbf{x}_t^{(n)}))^2/(2\sigma(\mathbf{x}_t^{(n)})^2)]$
%             \State Compute other costs
%         \EndFor
%         \State Evaluate total cost $\mathbb{J}^{(n)}$
%     \EndFor
%     \State Update $\mathbf{u}_{0:T-1}$ using MPPI weighted averaging
% \EndFor
% \State $\textbf{Output:}$ optimal control $\mathbf{u}_0^*$
% \end{algorithmic}
% \end{algorithm}
% \endgroup

\begin{algorithm}[t]
\small
\caption{MPPI with Motion-aware Traversability (MAT)}
\label{alg:mppi_mat}
\begin{algorithmic}[1]
\State \textbf{Input:} initial state $\mathbf{s}_0$, $\mathbf{u}_{0:T-1}$, height map $\mathbf{H}$, goal $\mathbf g$
\State Predict the MAT map $(A,\mu,\sigma) \leftarrow g_\theta(\mathbf{H})$
\For{iteration $k=1$ to $K$}
  \For{each sampled control perturbation $\delta\mathbf{u}^{(n)}_{0:T-1}$}
    \For{$t=1$ to $T-1$}
      \State $\mathbf{s}^{(n)}_t \leftarrow F(\mathbf{s}^{(n)}_{t-1},\mathbf{u}^{(n)}_{t-1})$,
             $(\mathbf{x}^{(n)}_t,v^{(n)}_t)\!\leftarrow\!\mathbf{s}^{(n)}_t$
      \State Accumulate cost $T(\mathbf{x}^{(n)}_t,v^{(n)}_t)$ and other terms
    \EndFor
    \State Compute trajectory cost $\mathbb{J}^{(n)}$
  \EndFor
  \State Update $\mathbf{u}_{0:T-1}$ via MPPI weighted averaging
\EndFor
\State \textbf{Output:} optimal control $\mathbf{u}^*_0$
\end{algorithmic}
\end{algorithm}

\begingroup
\renewcommand{\arraystretch}{0.89}
\begin{table*}[t]
  \centering
  \caption{Quantitative comparison between MAT and the baselines in the obstacle traversal task. For each metric, the average value (\textbf{Avg.}) across three velocity levels (induced by $T_1$, $T_2$, $T_3$) is reported. In each comparison, the better result is in bold.}
  % \vspace{-2mm}
  \begin{adjustbox}{width=\textwidth}\
  \newcolumntype{G}{>{\columncolor{baselinegray}}c}
  \begin{tabular}{c l *{3}{c} G *{3}{c} G *{3}{c} G}
    \toprule
    \multirow{2}{*}{\makecell{Terrain\\Geometry}} &
    \multirow{2}{*}{Method} &
    \multicolumn{4}{c}{Detour Distance (m) $\downarrow$} &
    \multicolumn{4}{c}{Traversal Time (s) $\downarrow$} &
    \multicolumn{4}{c}{Energy Consumption $\downarrow$} \\
    \cmidrule(lr){3-6} \cmidrule(lr){7-10} \cmidrule(lr){11-14}
    & & $T_1$ & $T_2$ & $T_3$ & \textbf{Avg.} & $T_1$ & $T_2$ & $T_3$ & \textbf{Avg.} & $T_1$ & $T_2$ & $T_3$ & \textbf{Avg.} \\
    \midrule

    % ----- Short Ditch -----
    % \rowcolor{baselinegray}
    \multirow{4}{*}{\cellcolor{white} Short Ditch}
    & PO-Trav 
    & 2.43 & 2.72 & 2.54 & 2.56
    & 18.49 & 10.89 & 8.56 & 12.65
    & 0.78 & \textbf{0.72} & 0.96 & 0.82\\
    & PhysORD \cite{zhao2024physord} 
    & 2.35 & 2.75 & 2.17 & 2.42
    & \textbf{17.96} & 12.64 & 8.54 & 13.05
    & 0.77 & 0.74 & 1.00 & 0.84 \\
    & AnyNav \cite{fu2025anynav} 
    & 2.60 & 2.46 & 2.49 & 2.52
    & 21.58 & 12.16 & 8.65 & 14.13
    & 0.97 & 0.81 & 1.01 & 0.93 \\
    & Ours
    & \textbf{2.30} & \textbf{0.06} & \textbf{0.05} & \textbf{0.81}
    & 18.25 & \textbf{9.21} & \textbf{8.22} & \textbf{11.90}
    & \textbf{0.77} & 0.87 & \textbf{0.78}  & \textbf{0.81} \\

    \midrule
    % ----- Long Ditch -----
    % \rowcolor{baselinegray}
    \multirow{4}{*}{\cellcolor{white} Long Ditch} 
    & PO-Trav 
    & 2.74 & 3.17 & 3.77 & 3.22
    & \textbf{16.87} & 13.66 & 6.92 & 12.49
    & 0.91 & 0.83 & 1.07 & 0.94 \\
    & PhysORD \cite{zhao2024physord} 
    & 2.39 & 3.27 & 3.91 & 3.19
    & 22.66 & 13.13 & 7.12 & 14.30
    & 0.81 & 0.88 & 1.09 & 0.93 \\
    & AnyNav \cite{fu2025anynav} 
    & 2.51 & \textbf{3.10} & 4.01 & 3.21
    & 19.76 & 12.80 & 8.49 & 13.69
    & 0.83 & \textbf{0.79} & 1.14 & 0.92
     \\
    & Ours 
    & \textbf{0.01} & 3.62 & \textbf{0.12} & \textbf{1.25}
    & 17.07 & \textbf{12.79} & \textbf{5.97} & \textbf{11.94}
    & \textbf{0.74} & 0.87 & \textbf{1.06} & \textbf{0.89} \\

    \midrule
    % ----- Bump -----
    % \rowcolor{baselinegray}
    \multirow{4}{*}{\cellcolor{white} Bump} 
    & PO-Trav 
    & 2.52 & 2.04 & 2.94 & 2.50
    & 17.91 & 16.29 & 7.75 & 13.98
    & 0.87 & 0.92 & 1.07 & 0.95 \\
    & PhysORD \cite{zhao2024physord} 
    & 2.98 & 2.17 & 3.35 & 2.83
    & 20.15 & 14.02 & 7.88 & 14.02
    & 0.85 & 0.79 & 1.05 & 0.90 \\
    & AnyNav \cite{fu2025anynav} 
    & 2.49 & 2.45 & 2.93 & 2.62
    & 19.35 & 14.79 & \textbf{7.29} & 13.81
    & 0.80 & 0.84 & 1.03 & 0.89 \\
    & Ours 
    & \textbf{0.07} & \textbf{0.06} & \textbf{0.15} & \textbf{0.09}
    & \textbf{17.45} & \textbf{13.71} & 8.41 & \textbf{13.19}
    & \textbf{0.79} & \textbf{0.76} & \textbf{0.98} & \textbf{0.84}
      \\

    \bottomrule
  \end{tabular}
  \end{adjustbox}
  \label{tab:sim_traverse}
  % \vspace{-2mm}
\end{table*}
\endgroup

\begin{figure*}[ht]
    \centering
    \includegraphics[width=\linewidth]{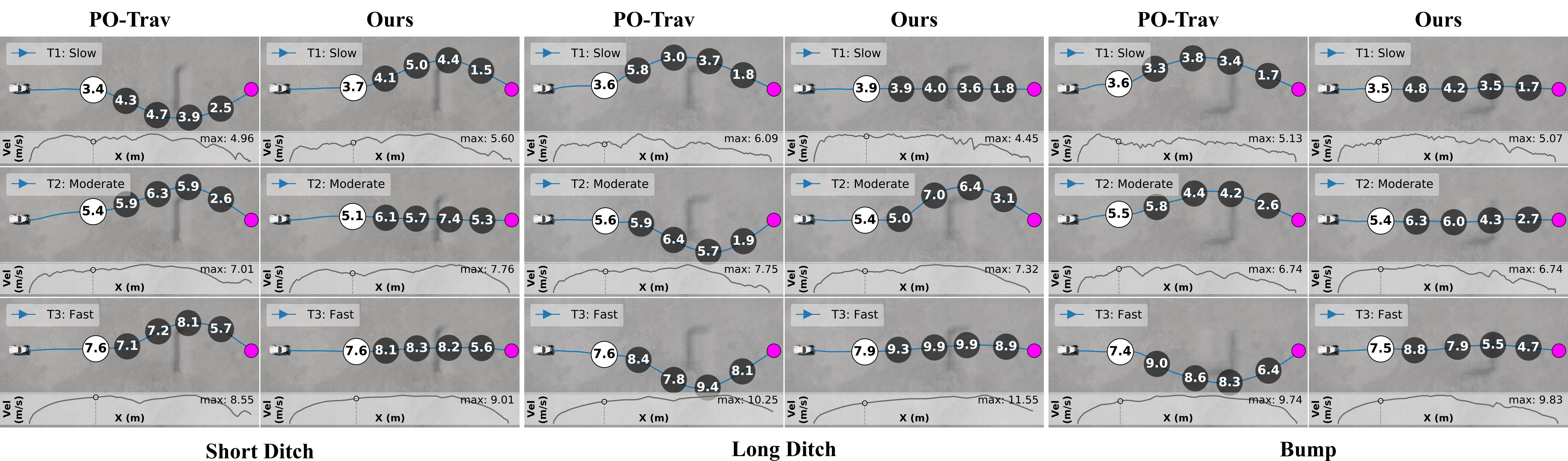}
    % \caption{Qualitative.}
    % \vspace{-2mm}
    \caption{Qualitative comparison between MAT and the PO-Trav in the obstacle traversal task. For each method, the vehicle trajectories with annotated velocities are shown for three obstacle types and three operating speeds. The velocity at which the obstacle first enters perception range is highlighted as a reference, and the full velocity profile is plotted for each test.}
    % \vspace{-2mm}
    \label{fig:sim_traversal}
\end{figure*}

\begin{figure*}[t]
    \centering
    \includegraphics[width=\linewidth]{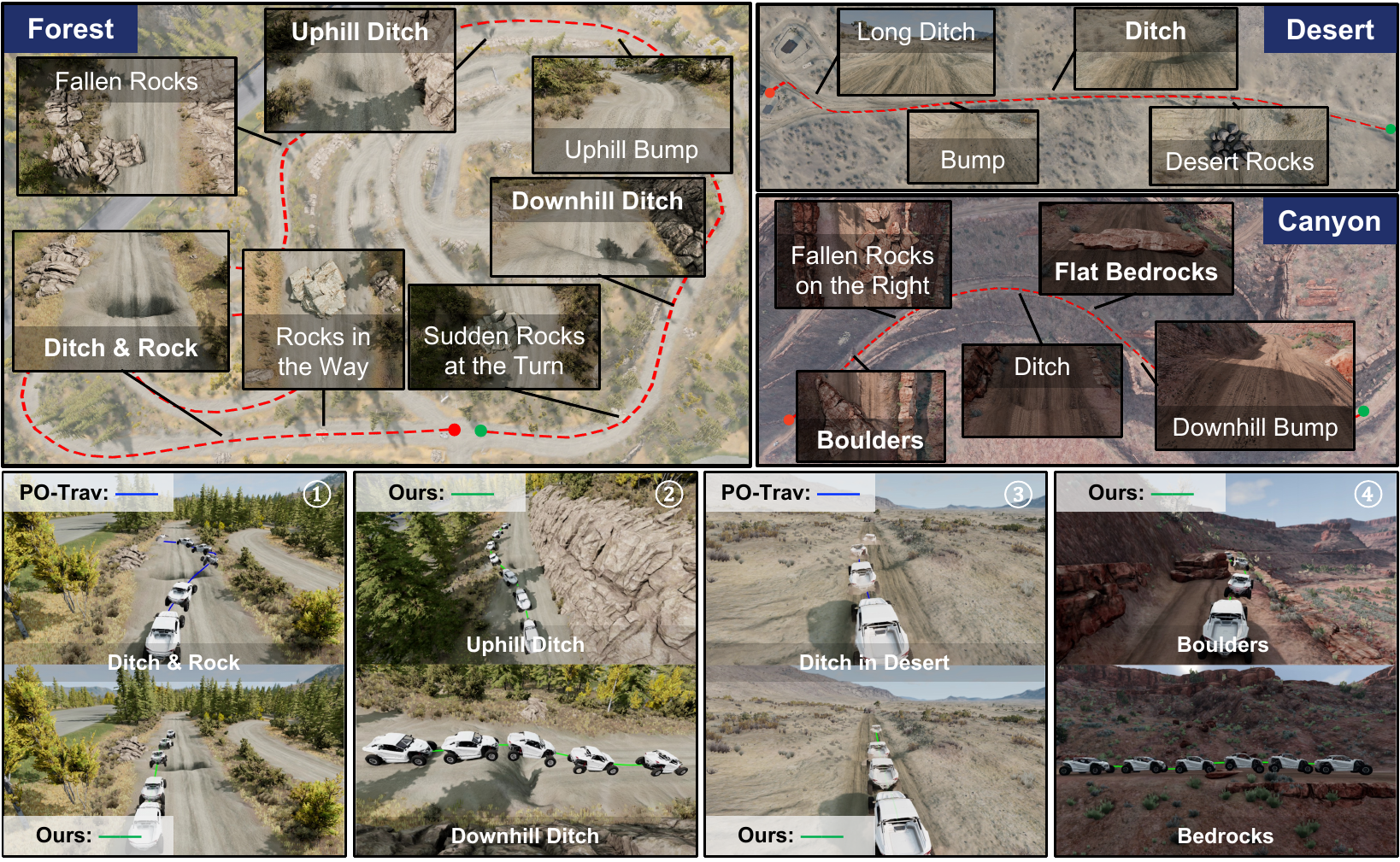}
    % \vspace{-6mm}
    \caption{Simulation maps and qualitative results for the long-range navigation task. The top row shows three modified maps with introduced obstacles blocking the routes. 
    The bottom row presents vehicle trajectories under MAT and the PO-Trav that exhibit distinct behavioral differences,
    with the corresponding obstacles highlighted in bold on the maps.}
    % This illustrates the obstacle (ditch, rock, bump) that we add at different map locations: (uphill, downhill, corners) on forest, canyon, and desert maps.} 
    % \vspace{-2mm}
    \label{fig:sim_nav}
\end{figure*}

\section{Experiments}
\label{sec:experiments}

\subsection{Implementation Details}
% This section describes the experimental platforms, implementation details, and results obtained with our approach.
\subsubsection{Platforms}
We use both simulation and real-world setups for training and evaluation.
Simulation is performed in BeamNG.tech \cite{beamng_tech}, which provides high-fidelity soft-body physics for accurate vehicle dynamics and tire-terrain modeling.
Its World Editor allows custom terrain geometry, enabling data collection for aggressive maneuvers that would be too costly in the real world.
% It provides diverse off-road maps and a built-in World Editor that allows customization of terrain geometry. This fidelity and flexibility enable aggressive vehicle maneuvers on challenging terrains for data collection and evaluation, which is costly and risky in the real world.
The real-robot platform is a Losi 1/5-scale \textit{Racer} car equipped with a MID-360 Livox, an NVIDIA Jetson Nano, and an Arduino. The setup supports on-board sensing, computation, and execution in real-world navigation.

\subsubsection{Configurations}
% \label{exp:imple}
\label{exp:metrics}
The neural network is a lightweight U-Net with four encoder-decoder stages and channel widths of $[16, 32, 64, 128]$.
Its input is a LiDAR-based height map generated on a $0.1 \mathrm{m} \times 0.1\,\mathrm{m}$ grid within a $40\,\mathrm{m}$ sensing range.
This range balances perceptual coverage with the foresight required for high-speed traversal.
The model is trained in simulation using Adam \cite{kingma2015adam} and fine-tuned on real-world data with a reduced learning rate. To avoid significant hardware damage, real data is collected in a conservative velocity range of $0.5$ to $6.5\text{ m/s}$. 
The MPPI is implemented within the MPPI-Generic \cite{vlahov2024mppi} framework, a C++/CUDA header-only library for conducting stochastic optimal control. We implement the dynamics model and our traversability cost function that conditions the terrain cost on the sampled motion.

\subsubsection{Baselines}
We use a Position-Only Traversability (PO-Trav) that shares the same configuration as MAT but replaces the motion-aware terrain cost with a position-only formulation.
This design provides a fair comparison that focuses on the contribution of our motion-aware representation by avoiding inconsistencies from different traversability definitions. Meanwhile, it captures the common essence of prior methods \cite{datar2025traverse,seo2023scate, seo2024metaverse}, and directly integrating these works is often hindered by the lack of full implementation releases \cite{gasparino2024wayfaster} or the difference in target robot types (e.g., legged robots \cite{frey23fast}).
% which largely do not release source code.
For comparison within the Motion-enhanced Traversability category defined in Section \ref{sec:related_work}, we include AnyNav \cite{fu2025anynav} and PhysORD \cite{zhao2024physord} in the simulation.
Both methods go beyond traversability estimation by improving off-road vehicle dynamics models for navigation and provide simulation implementations, while other motion-enhanced approaches lack released code for direct comparison.

\subsubsection{Metrics}
To quantitatively assess off-road navigation, we use three metrics. 
Detour distance measures the path length relative to the straight-line distance between start and goal.
While the values are not directly comparable across different scenes, this metric effectively compares planning efficiency within the same scene, especially involving open areas.
% as this baseline distance is not always the optimal path
Traversal time measures the total time taken and reflects the ability to maintain safe and efficient speeds. Energy consumption, computed as integrated throttle effort, 
captures the control workload required to complete the trajectory and serves as a proxy for overall power efficiency. Together, these metrics capture the agility, safety, and efficiency of a navigation policy.

% The MPPI time horizon directly affects the average sampled command velocities and thus the overall operating speed.
\subsection{Simulation: Obstacle Traversal}
\label{exp:sim_traverse}
To evaluate the ability of MAT to handle challenging obstacles, we consider a task in which the vehicle must find a feasible motion plan from a fixed start to a goal across obstacles.
% In each trial, the vehicle navigates from a fixed start to a goal with one challenging terrain segment in between. 
To emulate real-world conditions where a vehicle may encounter different obstacles at varying speeds, we test across two dimensions: (1) terrain geometry, including a short ditch, a long ditch and a bump; and (2) operating velocity, controlled by setting different MPPI time horizons [$T_1,\, T_2,\, T_3$] = [5, 3, 2] seconds, corresponding to slow, moderate, and fast motion. Under this setup, the only difference between each comparison pair lies in whether the terrain cost is conditioned on velocity. 

\subsubsection{Quantitative Comparison}
% We measure the metrics defined in Section \ref{exp:metrics}.
As summarized in Table \ref{tab:sim_traverse}, MAT reduces the detour distance by an average of 75\% across all obstacles compared to PO-Trav. 
MAT also achieves lower traversal time and energy consumption than AnyNav and PhysORD, as their dynamics models remain insufficient to model the coupled effects of terrain geometry and velocity.
Across different velocity levels, the gains in the averaged metrics primarily occur in scenarios where agile yet safe motion plans are possible and enabled by the MAT, as reflected in the per-velocity results and illustrated qualitatively next.

\subsubsection{Qualitative Comparison}
Fig. \ref{fig:sim_traversal} presents vehicle trajectories of MAT and PO-Trav, as AnyNav and PhysORD exhibit behaviors similar to PO-Trav.
Overall, MAT produces more adaptive and efficient motion plans than PO-Trav. For the short ditch, PO-Trav always avoids it due to its consistently high terrain cost. In contrast, MAT avoids the ditch when the vehicle moves slowly and accelerates to jump across it when sufficient speed can be achieved.
This flexibility arises because MAT first reasons about the velocities under current dynamics, then evaluates the terrain cost conditioned on those velocities. The Gaussian representation plays a significant role in enabling efficient terrain cost updates for new sampled velocities within the planner optimization loop.
This principle generalizes to other geometries. For the long ditch, which requires a higher speed to clear, MAT adapts its strategy: crawling over at low speed, avoiding at medium speed, and accelerating to jump at high speed. For the bump, which is only traversable at low velocity, MAT chooses to decelerate before crossing. These behaviors emerge from different Gaussian profiles $(A,\mu,\sigma)$ predicted from perception. In contrast, PO-Trav exhibits uniform avoidance due to the underlying limitation of position-only terrain cost, resulting in suboptimal plans. 
% \vspace{-3mm}
% Such plans may be safe in simple settings, but could be problematic when no clear detour exists or time is critical.

\begin{figure*}[t]
    \centering
    \includegraphics[width=\linewidth]{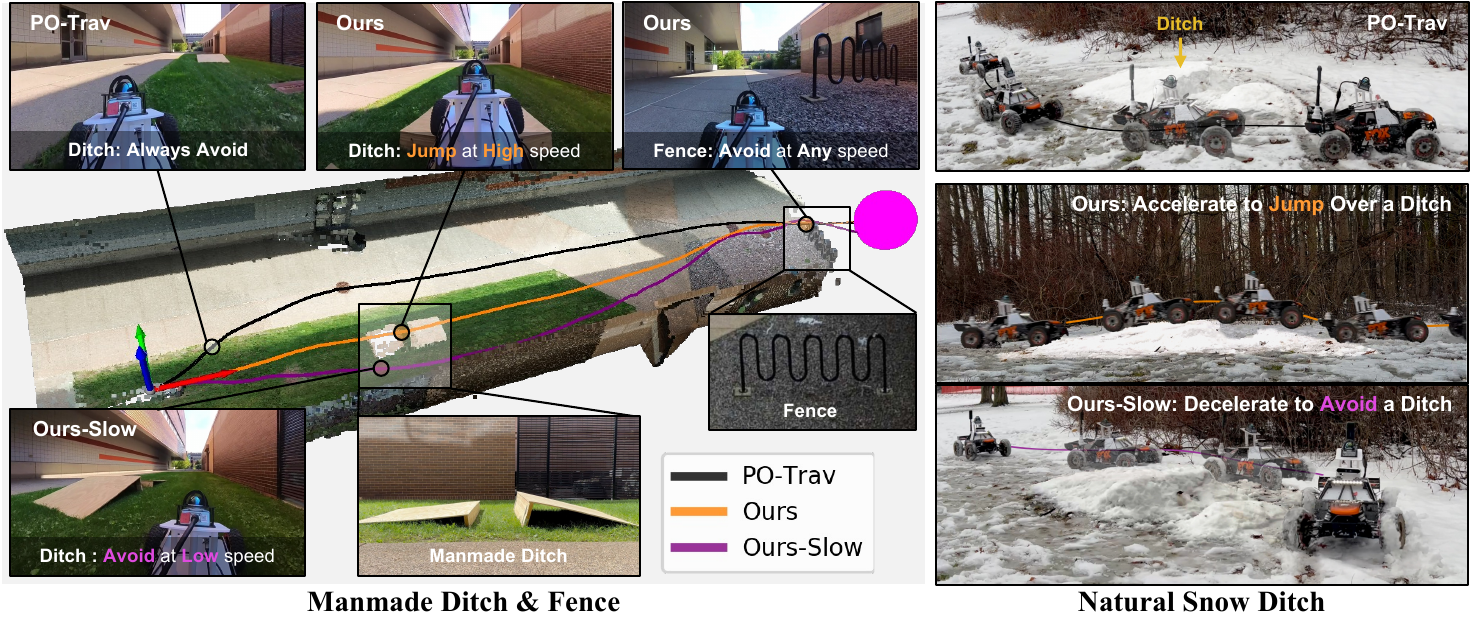}
    \vspace{-5mm}
    \caption{Real-world ditch traversal strategies in manmade (left) and natural (right) environments. While PO-Trav is consistently conservative, MAT demonstrates velocity-dependent reasoning by jumping (Ours) or avoiding (Ours-Slow). Simultaneously, MAT identifies the fence as universally non-traversable. Overlaid trajectories illustrate these distinct motion strategies.}
    % \vspace{-2mm}
    \label{fig:real_nav}
\end{figure*}

\begin{figure*}[ht]
    \centering
    \includegraphics[width=\linewidth]{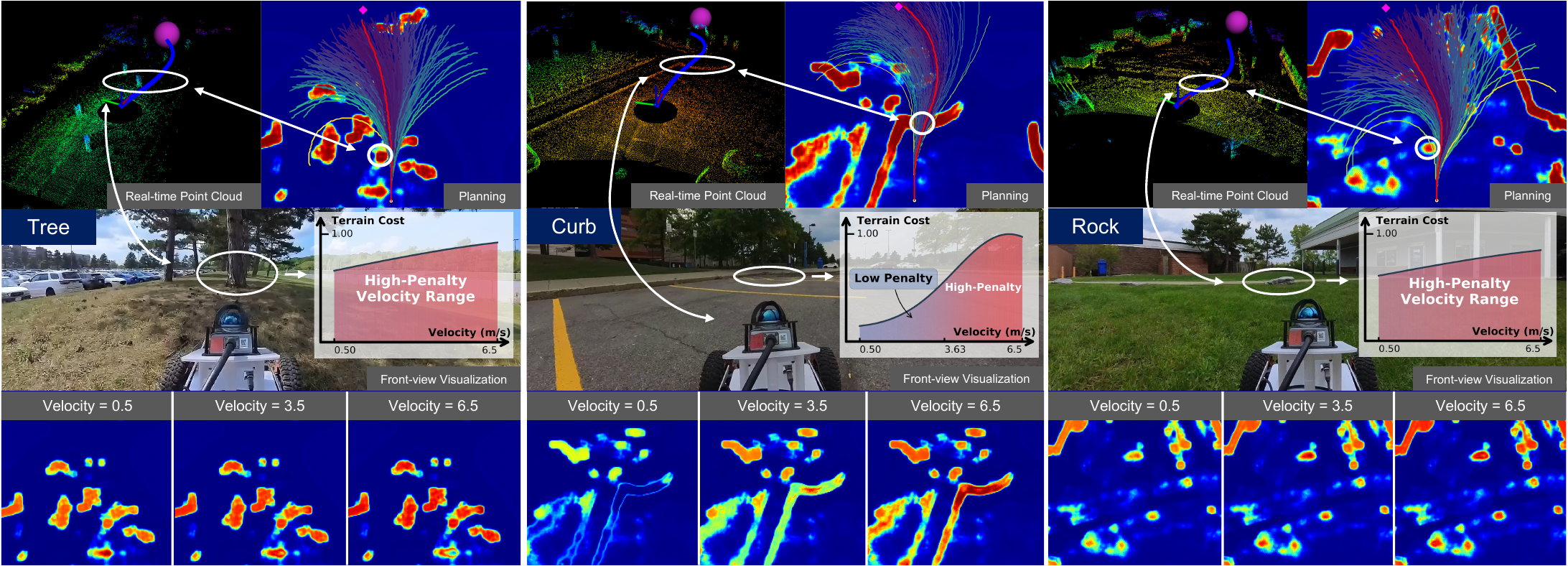}
    \vspace{-2mm}
    % \caption{Natural obstacles encountered in real-world navigation: tree, curb, and rock. The top row shows LiDAR points and sampled MPPI trajectories, with costs colored from blue (low) to yellow (high) and the local goal in magenta. The bottom row presents third-person front views of the obstacles and the predicted velocity-dependent terrain cost functions.} 
    \caption{Natural obstacles encountered in real-world navigation: tree, curb, and rock. Top: LiDAR points, sampled MPPI trajectories (where path costs are colored from blue/low to yellow/high and the goal is in magenta), and the predicted peak penalty map ($A(\mathbf x)$, defined in Eq.~\ref{eq:terrain_cost}). Middle: The predicted velocity-dependent cost function for a selected obstacle pixel and the third-person front view of the obstacle. Bottom: The resulting traversability maps when all pixels are evaluated at three constant velocities ($0.5\text{ m/s}$, $3.5\text{ m/s}$, and $6.5\text{ m/s}$), with map costs colored from blue (low) to red (high).
    % The middle row also presents third-person front views of the obstacles.
    } 
    \vspace{-2mm}
    \label{fig:real_nav_natural}
\end{figure*}

% Together, the qualitative and quantitative results demonstrate that MAT enables more optimal motion strategies across different terrain geometries and operating speeds.

\subsection{Simulation: Long-range Navigation}
To evaluate MAT in realistic, large-scale scenarios, we conduct long-range navigation experiments in BeamNG, where unstructured obstacles are added to appear unexpectedly and block the nominal route. In all experiments, the vehicle follows global waypoints spaced about $12\,\mathrm{m}$ apart, which do not account for obstacles or vehicle dynamics. 
Thus, the vehicle must rapidly adjust speed and trajectory to maintain safety.
% whenever new hazards arise.

\subsubsection{Navigation with Emerging Obstacles}
We first evaluate MAT on a $1084\,\mathrm{m}$ forest map. To increase complexity, we add obstacles such as ditches, rocks, and bumps across uphill, downhill, and curved regions (see Fig. \ref{fig:sim_nav}: Forest). 
The vehicle must autonomously reach the final goal by continually planning motions toward each waypoint while handling encountered obstacles in real time.
The resulting trajectories are shown in the numbered comparisons in Fig. \ref{fig:sim_nav}.
Compared with PO-Trav, MAT enables agile motion strategies, such as jumping a ditch in comparison (1). 
Moreover, MAT adapts its behaviors to the vehicle's motion dynamics. For instance, comparison (2) illustrates two contrasting scenarios: MAT avoids a ditch when approaching uphill with insufficient speed, but accelerates to jump across a downhill ditch when the momentum gained during descent makes the maneuver feasible. 
This experiment showcases MAT’s ability to account for speed-dependent traversability and respond effectively to emerging obstacles during long-range forest navigation.

\subsubsection{Generalization to Unseen Maps}
To assess generalization to unseen environments, we create two additional maps modified from the Utah Canyon ($638\,\mathrm{m}$) and Desert ($634\,\mathrm{m}$) environments. These maps introduce new terrain geometries, such as boulders and bedrocks, and distinct vehicle-terrain interactions, including sand surfaces.
Across both unseen environments, MAT preserves its agile and adaptive behavior. It reproduces the ditch-handling strategies observed in the forest experiment, as illustrated in comparison (3).
MAT also responds appropriately to new obstacle types, as shown in comparison (4): it safely avoids untraversable boulders and rocks and decelerates to crawl over flatter bedrock surfaces.

Together, comparisons (1) through (4) highlight MAT’s ability to rapidly and robustly adapt its motion strategy to both vehicle dynamics and unstructured terrain geometries.

\subsection{Real-world: Ditch Traversal}
To evaluate the real-world performance of MAT, we deploy our system on the \textit{Racer} in two ditch traversal experiments: a manmade environment and a natural field. 
Both scenarios involve obstacles with significant geometric challenges that are typically considered untraversable by prior methods, forcing traditional planners to take long detours. We compare MAT with PO-Trav, as AnyNav and PhysORD fail to produce stable results when deployed on \textit{Racer} for these scenarios.

% lack real-world implementation code, and
% rely on neural-based dynamics that are computationally heavy on edge hardware.

\paragraph{Manmade Ditch \& Fence}
In the manmade setting, the vehicle navigates toward a goal blocked by a ditch and a tall metal fence. We visualize the resulting trajectories in Fig.~\ref{fig:real_nav}, where the environment is reconstructed using COLMAP \cite{schoenberger2016sfm, schoenberger2016mvs}. The PO-Trav consistently avoids both obstacles due to the high terrain cost, leading to substantial detours. In contrast, MAT identifies that the ditch is traversable at high speed and accelerates to jump across it, while correctly recognizing the fence as untraversable at all velocities. When using the same MAT model with planner parameters adjusted to simulate a slow-moving vehicle (“Ours-Slow”), MAT correctly identifies that it cannot reach the velocity required for a safe jump and shifts its strategy to avoid the ditch. This highlights its ability to perform robust reasoning under current system dynamics.

\begin{table}[h]
% \vspace{-2mm}
\caption{Comparison in Real-world Ditch Traversal.}
\vspace{-2mm}
\centering
\small
\resizebox{\columnwidth}{!}{% % Forces the table to column width
\begin{tabular}{ll ccc}
\toprule
Scenario & Method & \begin{tabular}[c]{@{}c@{}}{Detour}\\{Distance [m] $\downarrow$}\end{tabular} & \begin{tabular}[c]{@{}c@{}}{Traversal}\\{Time [s] $\downarrow$}\end{tabular} & \begin{tabular}[c]{@{}c@{}}{Energy}\\{Consumption $\downarrow$}\end{tabular} \\
\midrule
\multirow{2}{*}{\begin{tabular}[l]{@{}l@{}}Manmade \\ Ditch \& Fence\end{tabular}} & PO-Trav & 1.54 & 15.40 & 6.13 \\
                         & Ours & \textbf{0.49} & \textbf{13.51} & \textbf{5.19} \\
\cmidrule{1-5}
\multirow{2}{*}{\begin{tabular}[l]{@{}l@{}}Natural \\ Snow Ditch\end{tabular}} & PO-Trav & 1.58 & 5.01 & 1.95 \\
                         & Ours & \textbf{0.09} & \textbf{2.01} & \textbf{0.81} \\
\bottomrule
\end{tabular}
}
\label{table:compact_results}
% \vspace{-4mm}
\end{table}

\paragraph{Natural Snow Ditch}
In the natural scenario, the vehicle must traverse an uneven, snow-filled ditch. Trajectories are visualized using sequential vehicle overlays to illustrate the traversal history. As in the manmade case, the PO-Trav avoids the ditch entirely, whereas MAT exploits higher speeds to traverse it directly and avoids it when operating under low-speed conditions that make a safe jump infeasible.

Across both experiments, MAT quantitatively outperforms the PO-Trav across all metrics. 
For a $40\,\mathrm{m}$ navigation task in the manmade setting, MAT reduces detour distance by 68.2\% compared to the PO-Trav. In the natural ditch experiment, MAT completes the traversal in 40.1\% of the time required by the PO-Trav. MAT also achieves lower energy consumption due to more efficient motion plans in both scenarios.

\subsection{Real-world: Outdoor Navigation with Obstacles}
% \vspace{-1mm}
We evaluate our system during autonomous navigation among diverse natural obstacles in the outdoor environment.
Fig. \ref{fig:real_nav_natural} shows obstacles encountered, including trees, curbs, and rocks, and the perception, MAT map prediction, and planning results in our navigation system.
Each obstacle type exhibits distinct traversability-velocity characteristics: trees and rocks maintain high terrain costs across all speeds, while curbs have low costs at low speed but high costs at high speed.
The MAT map captures these differences through varying Gaussian parameters $(A, \mu, \sigma)$, enabling the MPPI planner to generate adaptive maneuvers: avoiding trees and rocks while decelerating to crawl over curbs.
These results demonstrate MAT’s effectiveness in handling varied obstacles outdoors.

\section{Conclusions and Future Work}
\label{sec:conclusion}
% \vspace{-1mm}
This paper introduced Motion-Aware Traversability (MAT), a new representation that conditions terrain cost on both terrain perception and vehicle motion for off-road navigation.
By modeling traversability as a velocity-dependent function with parameters learned from perception, MAT captures how motion feasibility varies across terrains and enables efficient cost evaluation during planning. 
Experiments in BeamNG simulation and real-world settings show that MAT improves adaptability and efficiency, enabling vehicles to plan agile yet safe maneuvers such as accelerating to jump across ditches, slowing on bumps and curbs, and consistently avoiding untraversable obstacles. 
While the Gaussian function enables a step forward from previous traversability maps, it remains a simplified assumption. 
Furthermore, a primary system failure arises from complete LiDAR occlusion, where the system becomes overly conservative, but a higher sensor placement would resolve this issue.
Future work will extend MAT beyond the Gaussian model and incorporate additional motion factors to enhance robustness and expressiveness.

\section*{Acknowledgements}

This work was partially funded by ONR award N00014-24-1-2003 and DARPA award HR00112490426. The views and conclusions contained in this document are those of the authors and should not be interpreted as representing the official policies, either expressed or implied, of ONR or DARPA. The authors also thank Yi Du and Zitong Zhan (University at Buffalo) for technical support and insightful discussions.

{
\small
\bibliographystyle{plainnat}
\bibliography{references}
}
\end{document}